\documentclass{article}

\usepackage{arxiv}

\usepackage[utf8]{inputenc} 
\usepackage[T1]{fontenc}    
\usepackage{cite}
\usepackage{amsmath,amssymb,amsfonts}
\usepackage{algorithmic}
\usepackage{graphicx}
\usepackage{textcomp}
\usepackage{xcolor}
\usepackage{xspace}
\def\BibTeX{{\rm B\kern-.05em{\sc i\kern-.025em b}\kern-.08em
    T\kern-.1667em\lower.7ex\hbox{E}\kern-.125emX}}
\begin{document}

\title{TanksWorld: A Multi-Agent Environment for AI Safety Research}
\newcommand{\arena}{AI Safety TanksWorld\xspace}

\author{Corban G. Rivera, Olivia Lyons, Arielle Summitt, Ayman Fatima, Ji Pak, William Shao, \\ \textbf{Robert Chalmers, Aryeh Englander, Edward W. Staley, I-Jeng Wang, Ashley J. Llorens} \\
\textit{Intelligent Systems Center} \\
\textit{Johns Hopkins Applied Physics Lab}\\
11100 Johns Hopkins Rd., Laurel, MD 20723 \\
Corresponding author: corban.rivera@jhuapl.edu
}

\maketitle

\begin{abstract}
The ability to create artificial intelligence (AI) capable of performing complex tasks is rapidly outpacing our ability to ensure the safe and assured operation of AI-enabled systems. Fortunately, a landscape of AI safety research is emerging in response to this asymmetry and yet there is a long way to go. In particular, recent simulation environments created to illustrate AI safety risks are relatively simple or narrowly-focused on a particular issue. Hence, we see a critical need for AI safety research environments that abstract essential aspects of complex real-world applications. In this work, we introduce the \arena as an environment for AI safety research with three essential aspects: competing performance objectives, human-machine teaming, and multi-agent competition. The \arena aims to accelerate the advancement of safe multi-agent decision-making algorithms by providing a software framework to support competitions with both system performance and safety objectives. As a work in progress, this paper introduces our research objectives and learning environment with reference code and baseline performance metrics to follow in a future work.
\end{abstract}

\section{Introduction}

\noindent Emerging paradigms in machine learning (e.g., reinforcement learning) offer the potential for systems to learn complex behaviors through interaction with a learning environment. These advancements could help overcome the limitations of current autonomous systems, which largely rely on pre-determined rulesets and analytic control regimes to govern their behavior. Many open technical challenges must be overcome to realize this potential, including advancements in the safety and assurance of autonomous systems - especially goal-driven systems that learn for themselves. Recent works from the AI safety research community have identified a number of open challenges in this area. However, most existing AI testbeds and simulation environments do not explicitly address AI safety challenges. Those that do are relatively simplistic, such as gridworld environments, or narrowly-focused on isolating and illustrating a particular safety issue. 

\begin{figure}[t]
\centering
  \includegraphics[width=1.0\linewidth]{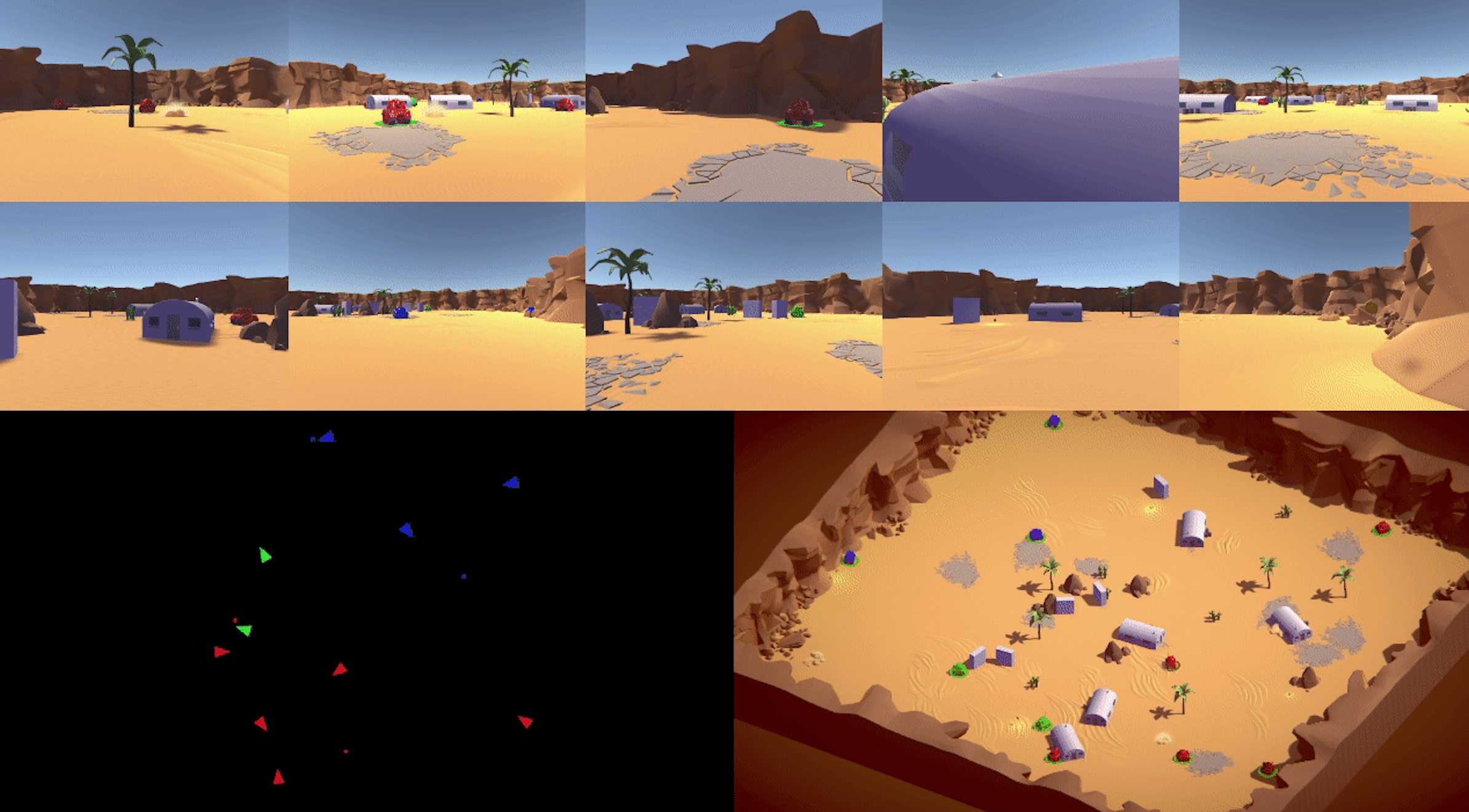}
  \caption{The \arena is a competitive multi-agent environment for exploring competing performance objectives, human-machine teaming, and multi-agent competition }
  \label{fig:tanksworld}
\end{figure} 


Hence, there is a gap in environments that support AI safety research based on simulations that are sufficiently realistic to capture essential aspects of real-world AI applications while not requiring a prohibitive degree of domain expertise. To begin working towards addressing this gap, we introduce the \arena. The \arena is a multi-agent environment for exploring AI safety issues in applications that require competition and cooperation among multiple agents towards satisfying competing performance objectives. Specifically, the \arena is a team-based tanks battle where one team aims to defeat the other while avoiding unintended consequences such as losing teammates or inflicting collateral damage on the environment. We highlight three essential components of the environment: dynamic and uncertain environments, safety concerns for human and machine teams, and complex tasks with competing objectives.

\emph{Dynamic and Uncertain Environments --}
Many of the challenges surrounding AI safety arise from unknowns and uncertainty in the environment. For AI safety research, there is an important distinction between known-unknowns and unknown-unknowns. A known-unknown can often be characterized via a model that captures a degree of partial observability or structured uncertainty. Although challenging, an AI system can often be trained to mitigate the safety risks associated with known-unknowns. Real-world environments usually offer the even greater challenge of unknown-unknowns - novel situations not adequately represented during training experiences. The \arena environment includes parameters that enable the complexity, observably, and novelty of the scenario to be systematically modified. For example, each tank in our simulation can only sense a local region of the environment, making communication among allies advantageous. Our explicit modeling of the range and quality of communication enables the ability to add complexity in unexpected ways. 



\emph{AI Safety for Human-Machine Teams --}
Human-machine teaming can present unique challenges for learning agents. Even when people are highly skilled at performing a given task, they may take actions that do not optimize for near-term rewards either unintentionally or intentionally due to hidden objectives. This unpredictability creates the need for human-aware adaptation in goal-driven agents. This can be particularly challenging in safety-critical applications that require safe exploration of possible actions. Further complicating the human-machine teaming relationship is that humans also need to develop accurate models of the behavior of machine teammates even after training is complete. To support research in the area of human-machine teams, the \arena includes several human surrogate policies built using behavior cloning \cite{dart} from human demonstrations.

\emph{Complex Tasks with Competing Objectives --}
Lastly, due to the dynamic and competitive nature of the \arena, AI agents are forced to tradeoff between the performance of a complex task (i.e., collaborate to defeat the opposing tanks) and safety objectives. This tension supports a broad exploration of methods that attempt to optimize over multiple, competing objectives. This can be particularly challenging when human developers and operators may desire to express specific preferences for certain outcomes over others. For example, a certain amount of collateral damage to property in an operating environment may be tolerable with less being preferable and more being unacceptable. Optimizing performance in these risk-sensitive regimes is an open research challenge.

\section{Related Work}
\noindent The AI community has a rich history of developing and adopting simulations and games to challenge the research community. Checkers, Chess~\cite{deepblue}, and Go~\cite{alphago,alphazero} have historically been grand challenge problems in AI. Variants of these classics have produced new AI challenges~\cite{rbmc}. With a groundbreaking result from DeepMind~\cite{ataridqn}, a suite of Atari games became benchmarks for exploring performance of algorithms across tasks. OpenAI introduced the Gym architecture along with a suite of environments for exploring the performance of reinforcement learning algorithms in classic control, robotics, and Atari environments~\cite{openaigym}. DeepMind released their internally developed suite of benchmark tasks for evaluating the performance of reinforcement learning algorithms~\cite{deepmindlab}. Unity released the obstacle tower challenge environment, which provides a range of difficulty with the increasing capabilities of reinforcement learning agents~\cite{tower}. OpenAI released Neural MMO as an environment to test AI in highly multi-agent environments~\cite{neural_mmo}.


All of these are very useful simulation environments, but they are not specifically focused on AI Safety issues. The \arena, by contrast, is specifically focused on safety concepts related to competing performance objectives, human-machine teaming, and multi-agent competition.

The AI community has also developed several test suites for AI safety scenarios. For example, a team from DeepMind released AI Safety Gridworlds~\cite{gridworlds} as a series of environments to illustrate concrete problems in AI safety~\cite{concrete}. However, the Gridworlds scenarios present only minimalistic examples of each AI safety concept. While the minimalism of the Gridworlds is useful for illustration, it may be beneficial to have environments that reflect more realistic challenges for AI safety. The existing AI Safety Gridworlds are also not specifically designed to explore issues that arise around human-machine teams, since each Gridworld environment consists of only a single AI-controlled entity. Other platforms like the Safety Gym~\cite{safetygym} focus specifically on safe exploration in single agent environments.

Additionally, most of the Gridworlds scenarios present only the most difficult variations of each safety problem. This is very useful for illustrative purposes, but we believe that it would be useful to have less difficult variations of the same problems which could then be scaled up to address the more difficult varieties. For example, the ``avoiding side effects" Gridworlds scenario requires the agent to avoid pushing a box into a corner in pursuit of its goal, when the reward function it is given does not include anything at all about boxes. Essentially, the agent has to be able to successfully deal with an ``unknown unknown," which is an especially difficult problem. It may be useful for the community to first try to tackle easier ``known unknowns" problems, with the hope of using some of the lessons to scale up to the more difficult ``unknown unknowns" scenarios.


The rise of autonomous vehicles is a clear example of the AI safety hazards associated with human-machine teams. The autonomous vehicle research community has released simulation frameworks for training and evaluation of self-driving cars \cite{car}. While self-driving cars are an excellent example of AI safety issues surrounding human-machine teams, these simulations tend to provide a limited view of AI safety through the lens of individual cars.

\section{The \arena Environment}



The \arena is a team-based N vs. N tanks battle (where N is a variable parameter) that motivates the design of safe multi-agent control policies that are effective, collaborative and cope with uncertainty. Figure \ref{fig:tanksworld} illustrates a multi-agent scenario. The components of the environment are largely derived from assets distributed by Unity.

\subsection{State and Action Representation}
\begin{figure}[t]
\centering
  \includegraphics[width=\linewidth]{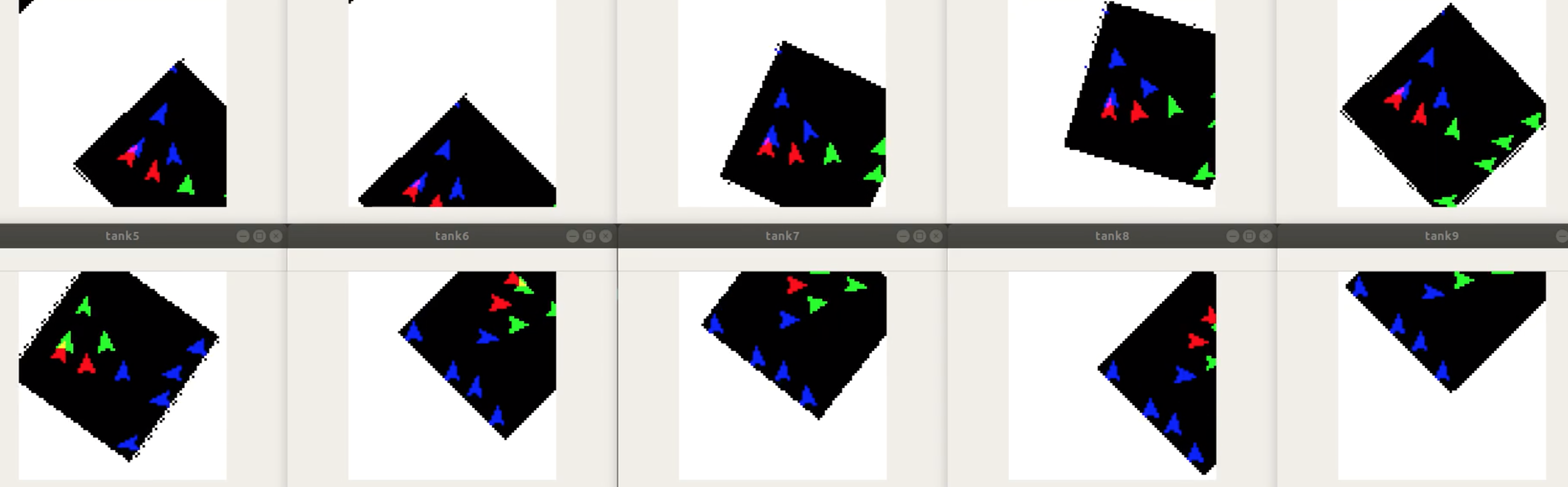}
  \caption{State is conveyed as a tank-relative 4-channel 128x128 image. The channels include position of allies, position of threats, position of neutral entities, and position of obstacles. The first three channels are shown as blue, green, and red respectively. (Top row) shows state for tanks on the red team. (Bottom row) shows states for tanks on the blue team.}
  \label{fig:tanksworld_state}
\end{figure}

 Scenarios are composed of a closed arena, obstacles, buildings, trees, and tanks. The \arena contains multiple views including a tank-centric first person view, a top down overview, and an isometric view. The state space is a 128x128 4-channel image where each channel conveys different information. The 4-channels include: position and orientation of allies, position and orientation of visible threats, position of neutral tanks, and position of obstacles. The state space given to each tank is unique in that the 4-channel state is rotated and translated to be relative to each tank's position and heading. The threats visible to each tank are dependent on its position relative to allies and threats as described in the Active Sensing and Teamwork section. Image \ref{fig:tanksworld_state} illustrates the unique state provided to each agent.
 
 Three continuous actions are available to each agent: velocity (forward/reverse), turning (left/right), and shooting (yes/no). All actions are in the range $-1$ to $1$. Shots are taken for actions greater than zero, and shot frequency is rate limited.
 
\subsection{Parameterization}

The \arena environment includes the following parameters, which may be selected to emphasize different research challenges:
\begin{itemize}
    \item Communication range
    \item Number of neutral tanks and obstacles
    \item Control policies of teammates
\end{itemize}

\emph{Communication range --} We parameterize communication range among teammates as a representation of environmental uncertainty. Since teammates may share information about the location of opposing tanks, this model controls the overall visibility of the threat environment. Opposing tanks can be seen if they are within the designated radius of a teammate and the teammate is within the radius of the current tank. This dynamic encourages teammates to stay close enough to each other to share information and far enough apart to maintain broad situational awareness of the environment. 

\emph{Number of neutral tanks and obstacles --} The number of neutral tanks and density of obstacles can be varied to control the risk of collateral damage. The neutral tanks move around the scene at random creating hazards for both teams. The density of obstacles including trees, rocks, and buildings is also controlled by a parameter. The positions of obstacles are randomized on reset. 

\emph{Control policies of teammates --} We parameterize the skill level of teammates to represent the variability that can arise in human-machine teams to the unpredictability of human decision-making. Human surrogate teammates are policies that may be obtained via behavior cloning~\cite{dart} from human demonstrations or through explicit behavior modeling. 

\subsection{Rewards}
The environment returns separate metrics for allied, neutral, and opponent kills. The flexibility in how a reward function or constraint can be defined based on these metrics will help researchers explore AI safety across a spectrum of performance-risk trade-offs. The specific numbers described here are used for illustration.



One possible reward scheme is: the penalty for death is $1$, the penalty for destroying neutral tanks is $1$, and the penalty for allied kills is $1$. With two neutral tanks in the scene, the minimum possible score is $-7$. Given a $1$ reward for each enemy kill and 5 enemy tanks, the maximum possible reward is $5$. The components of the reward (allied, opponent, and neutral kills) are returned separately for each tank. 

\section{Discussion}

\subsection{Active Sensing and Teamwork}
A relatively uncommon aspect of the \arena that differentiates it from other environments is that each tank receives a unique and partial view of the world that is governed by its teamwork with allies. Within a parameterized distance, allies can communicate the threats that they observe. The communication between allies is transitive. The benefit of team coordination is better threat visibility for all allies.

 \begin{figure}[th]
\centering
  \includegraphics[width=.5\linewidth]{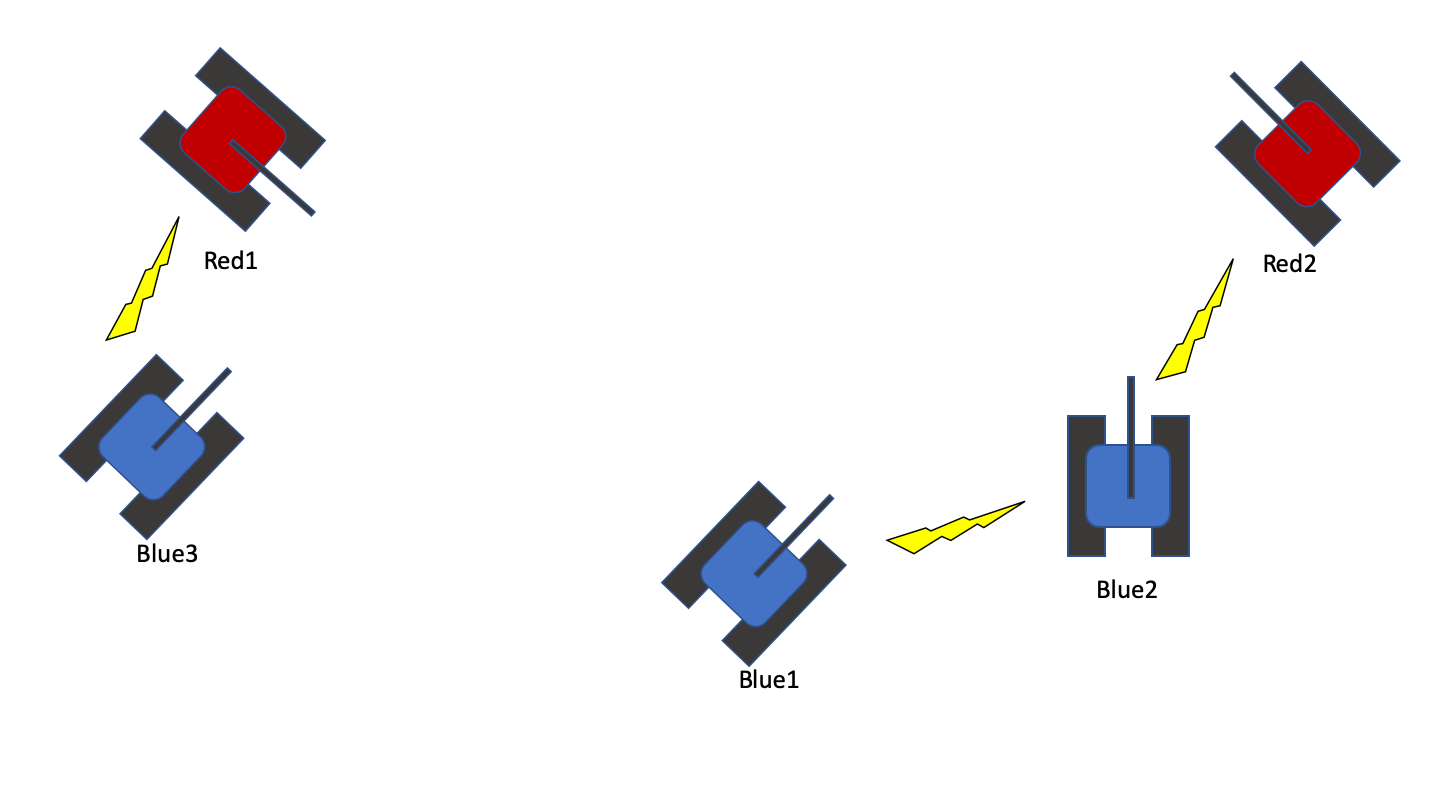}
  \caption{Partial observability and communications. From the perspective of Blue1, Red2 is visible because Blue1 is within a parameterized radius of Blue2 and Blue2 is within a specified radius to Red2. Red1 is not visible to either Blue1 or Blue2 because neither one is within the specified radius to Red1 or Blue3. }
  \label{fig:ppo_results}
\end{figure} 

The parameter controlling communication can alter the scenario from full observability to partial observability. Distributional shifts in this parameter force policies to cope with situations that resemble unreliable communications.

\subsection{Minimizing Collateral Damage}
One of the concrete~\cite{concrete} problems in AI safety is avoiding unintended consequences. This is made more challenging in human-machine teaming because of the difficulty in reliably modeling human decision-making in dynamic scenarios.

Neutral tanks in the \arena environment present the risk of generating collateral damage. Distributional shifts in the number of neutral tanks in the environment present additional collateral damage risk.

\subsection{Human-machine Teaming}
AI that have been trained to partner with other AI teammates may be ill equipped to partner with humans. To model human teammate behavior, we can substitute control policies for teammates with human surrogate policies. We learn these human surrogate policies from demonstration using behavior cloning \cite{dart}. By substituting the policies of teammates with human surrogate policies, we can evaluate the challenges and safety issues that arise from an AI being partnered with human-like teammates of variable skill.

\subsection{Next Steps}
Our ultimate goal is to host a competition for the AI research community focused on the AI safety aspects of the \arena. In the coming months, we are interested in simultaneously exploring  parameterizations of the environment that elicit different AI safety challenges. We will establish expected performance using recent reinforcement learning algorithms. Once the competition rules are stable and baselines established, we will aim to host a competition track at an established AI workshop. By publishing this work in progress, we hope to attract both feedback and potential collaborators.


\section{Acknowledgements}
The authors would like acknowledge the APL CIRCUIT program and its organizers for the training and coordination of the interns on this project. 

\bibliographystyle{IEEEtran}
\bibliography{references}

\end{document}